\documentclass[
]{ceurart}

\usepackage{listings}
\begin{document}

\copyrightyear{2025}
\copyrightclause{Copyright for this paper by its authors.
  Use permitted under Creative Commons License Attribution 4.0
  International (CC BY 4.0).}

\conference{In: R. Campos, A. Jorge, A. Jatowt, S. Bhatia, M. Litvak (eds.): Proceedings of the Text2Story'25 Workshop, Luca (Italy), 10-April-2025}

\title{Narrative Shift Detection: A Hybrid Approach of Dynamic Topic
Models and Large Language Models}


\author[1]{Kai-Robin Lange}[%
orcid=0000-0003-1172-9414,
email=kalange@statistik.tu-dortmund.de,
]
\cormark[1]
\author[2]{Tobias Schmidt}[%
orcid=0000-0001-7330-4943,
email=tobias3.schmidt@tu-dortmund.de,
]
\author[3]{Matthias Reccius}[%
orcid=0000-0002-0716-0432,
email=Matthias.Reccius@ruhr-uni-bochum.de,
]
\author[2]{Henrik Müller}[%
orcid=0009-0004-1747-3303,
email=henrik.mueller@tu-dortmund.de,
]
\author[3]{Michael Roos}[%
orcid=0000-0002-5465-9893,
email=michael.roos@ruhr-uni-bochum.de,
]
\author[1]{Carsten Jentsch}[%
orcid=0000-0001-7824-1697,
email=jentsch@statistik.tu-dortmund.de,
]

\address[1]{Department of Statistics, TU Dortmund University, 44221 Dortmund, Germany}
\address[2]{Institute of Journalism, TU Dortmund University, 44221 Dortmund, Germany}
\address[3]{Faculty of Management and Economics, Ruhr University Bochum, 44780 Bochum, Germany}

\cortext[1]{Corresponding author.}

\begin{abstract}
  With rapidly evolving media narratives, it has become increasingly critical to not just extract narratives from a given corpus but rather investigate, how they develop over time. While popular narrative extraction methods such as Large Language Models do well in capturing typical narrative elements or even the complex structure of a narrative, applying them to an entire corpus comes with obstacles, such as a high financial or computational cost. We propose a combination of the language understanding capabilities of Large Language Models with the large scale applicability of topic models to dynamically model narrative shifts across time using the Narrative Policy Framework. We apply a topic model and a corresponding change point detection method to find changes that concern a specific topic of interest. Using this model, we filter our corpus for documents that are particularly representative of that change and feed them into a Large Language Model that interprets the change that happened in an automated fashion and distinguishes between content and narrative shifts. We employ our pipeline on a corpus of The Wall Street Journal news paper articles from 2009 to 2023. Our findings indicate that a Large Language Model can efficiently extract a narrative shift if one exists at a given point in time, but does not perform as well when having to decide whether a shift in content or a narrative shift took place.
\end{abstract}


\maketitle

\section{Introduction}
With the rise of populism in western democracies, it has become increasingly important to evaluate narratives in politics, economics and other areas of interest that shape our society. Broadly speaking, narratives are linguistic constructs that boil complex connections between events down to an explainable form. Such narratives do not have to fall in line with facts and have become increasingly important in many areas of our society as they can act as a substitute for fact-based decision making. Consequently, methods to extract and analyze such narratives from large corpora have garnered the interest of many researchers. With both constantly changing and also conflicting narratives spread by entities such as policy makers, news outlets or social media personalities, we consider it especially important to observe, how a narrative develops over time instead of just globally extracting narratives from a corpus that spans over a long time period.

Along with the development and improvement of Natural Language Processing (NLP) methods, narrative extraction methods have also improved. The latest big development in NLP has brought a push towards more reliable human-like narrative extraction: the language understanding capabilities of Large Language Models (LLMs) such as ChatGPT \cite{openai_gpt-4_2024}, Llama \cite{dubey_llama_2024}, and others show the ability to detect narratives with increasingly complex definitions and thus blur the lines between qualitative and quantitative analyses. Given billions of parameters in the neural networks these models are based on, they can answer questions about almost any input document and the narratives contained within, but this enormous size also comes with drawbacks. Similarly to how expensive it is to pay experts of a certain field of interest to annotate texts, using LLMs for large scale corpora is not feasible for many researchers due to their computational demand or the financial cost of commercial models. Additionally, their size does not allow users to fully train the models on just a single corpus from scratch. Thus, many methods that use training from scratch to model temporal developments in a corpus, such as many dynamic topic models and diachronic word embedding models \citep{kutuzov_diachronic_2018}, cannot be directly transferred to LLMs, leaving fewer options for temporal narrative change detection.

We propose a pipeline that combines the best qualities of both approaches, using dynamic topic models and LLMs. Leveraging the approach of \cite{TopicalChanges}, we use a bootstrap-based topical change detection on the topics resulting from the dynamic topic model \enquote{RollingLDA} \cite{rieger_rollinglda_2021}. While this model succeeds in providing us with change points that are based on differences in the word count vectors of individual topics, it does not give us any intricate details about the changes themselves except for the topic of the change, the time chunk in which it happened and some key words that are responsible for the detection of the change. Furthermore, when a change is detected, it is not always certain whether it actually signifies a genuine shift in the narrative or in another dimension of the discourse, such as the factual content or the contextual framework provided. We can however use this method to narrow a large corpus down to a small curated number of documents that are suspected to contain information about some discoursive shift. We do this by filtering the documents within the time period in which the change occurred, given the information we are provided about the change. We then use the language understanding capabilities of an LLM by processing these documents to explain the topical change that occurred and to decide whether the change signifies a narrative shift or not. We additionally provide the LLM with information gathered by the topic model to put the change into context. To accurately guide our identification of shifts in narratives, we use the Narrative Policy Framework (NPF) \cite{jones2010narrative, shanahan2018narrative}, an analytical approach from political science. Our findings indicate that the LLM performs well when explaining a narrative shift, if one exists, but hallucinates when judging whether a detected change is a narrative shift or not, claiming most content shifts to also be narrative shifts.

We evaluate our model on a corpus of news articles of The Wall Street Journal ranging from 2009 to 2023. As the documents are copyright-protected, we do not use a commercial LLM in the cloud, but rather a local instance of the open source LLM Llama 3.1 8B \citep{dubey_llama_2024}.

\section{Related Work}
In our research, we focus on media narratives stemming from news articles with a business and finance focus. Thus, we base our definition of narratives on the existing literature on narratives from research in the field of economics and political economics.

\subsection{Defining Narratives}

The study of narratives has recently gained traction in economic research, though scholars have yet to converge on a single, universally accepted definition of the concept. \citet{shiller2017narrative}, as an early and influential example, characterizes economic narratives as \enquote{stories that offer interpretations of economic events, or morals, or hints of theories about the economy}, thus providing a rather vague definition that leaves ample room for interpretation. A more formal modeling strategy was pioneered by \citet{eliaz2020model}, who adapt concepts from the literature on Bayesian networks \cite{Pearl2009}. 
The authors highlight the causal connections among events and economic variables as sufficient for shaping people’s economic and political beliefs. While analytically appealing, this conceptualization sets aside value judgments, a crucial aspect of narratives. We hold that these normative implications ground the impact of political and economic narratives by motivating individuals and groups to act on their beliefs \cite{roos2024narratives, Shenhav2006, Eliaz2024a}.

To empirically capture narratives, including their ideological valence, the Narrative Policy Framework (NPF) offers an alternative lens \cite{jones2010narrative, shanahan2018narrative, schlaufer2022narrative}. The NPF distinguishes narratives by their content and form, systemizing the latter through four elements: a setting, certain characters, a plot and a moral of the story. 
The setting describes the scenery in which the narrative takes effect, such as a presidential election, a military conflict or a time of high inflation. The characters of a narrative can be persons or organizations, but also other entities take actions in the narrative including even non-sentient entities such as a spreading virus. The plot establishes relationships between the characters in space and time. Lastly, the moral of the story acts as a \enquote{takeaway} that often includes implicit or explicit calls to action.
With these components, the NPF highlights both the role of ideological charge and the centrality of causal attributions, particularly by emphasizing that identifying \enquote{who is to blame for the problem} is an essential part of every narrative \cite{crow2018narratives}.

Similarly, \citet{muller2018wert} propose a narrative definition that is based on six key elements. They build their framework around the theory of media frames \cite{entman1993framing}, proposing that narratives provide dynamic, evolving depictions of events. According to this view, a media narrative comprises one or more media frames (all of which are built around four key elements, see \cite{entman1993framing}) combined with protagonists (e.g., individuals and institutions) and events arranged chronologically and often presented as causally linked. 


For a more comprehensive overview of narrative definitions across disciplines, see the overview papers by \citet{roos2024narratives} or \citet{santana2023survey}, among others.

Building upon foundational works in narrative theory, our research focuses on detecting narrative structures and their evolution in large text corpora. To achieve this, we draw on methodologies that have proven effective in identifying thematic shifts or \textit{change points} within textual data. These change points often coincide with shifts in journalistic focus, and we hypothesize that such shifts frequently reflect underlying narrative changes. By leveraging established methods that combine topic modeling with change point detection, we aim to capture these transitions, providing valuable insights into how narratives develop and evolve over time.

\subsection{Extracting Narratives from Text}

Following the identification of discursive change points, the second critical step in our approach involves the automated analysis of the documents associated with these transitions. To date, there is no widely established (language) model specifically optimized for narrative extraction in this context. Instead, a diverse range of methods has emerged, each attempting to identify narratives or their components through diverse techniques ranging from word-count-based analyses \cite{NEN} to Large Language Model-based methods \cite{gueta2024can}.

A detailed review of existing NLP-methods to identify systematic parts of narratives is outlined by \citet{santana2023survey}. The authors focus on identifying key components such as events, participants, and temporal and spatial data, and linking these components to form coherent narratives. The methodology discussed includes part-of-speech tagging, event extraction, semantic role labeling, and entity linking, among others. However, this approach using classical NLP-tools struggles with challenges such as narrative complexity and cross-document narratives. 

A similar approach, focusing on political and economic narratives, is proposed by \citet{ash2024relatio}. Their methodology \textit{RELATIO} employs semantic role labeling to identify key narrative components such as agents, actions, and patients within sentences, which culminates in the production of interpretable narrative statements. While their approach effectively identifies simple narrative building blocks, it falls short in detecting and extracting more complex narratives that integrate causality and sense-making. \citet{lange_towards_2022} advanced this approach by enhancing the existing RELATIO method with additional pre- and post-processing steps. By combining multiple RELATIO-extracted narrative blocks, the authors were able to link related statements and extract complex narrative structures that better align with causality-based definitions of narratives. At the same time, they emphasize that the increased complexity of their pipeline can amplify error cascades, where even minor changes in a longer input sequence may lead to significantly different results.

A recent study leveraging large language models (LLMs) for narrative extraction is presented by \citet{gueta2024can}. The authors explore whether LLMs can effectively capture macroeconomic narratives from social media platforms like X (formerly Twitter). However, the study falls short of providing a robust definition of narratives, focusing instead on sentiment analysis and RELATIO-style statements, leaving key aspects of narrative complexity and causality underexplored.

Collectively, these studies demonstrate the evolving landscape of narrative extraction methodologies, highlighting the integration of advanced NLP techniques to unravel complex narrative structures. By employing a hybrid approach, we seek to address the limitations identified in previous studies, particularly concerning the temporal dynamics of narratives and their computational feasibility. Our proposed pipeline, which combines dynamic topic models and LLMs, aims to provide a more comprehensive understanding of narrative changes over time, thereby contributing to the broader discourse on narrative extraction and analysis.

The choice to utilize LLMs for annotating and categorizing economic narratives stems from their demonstrated ability to excel at complex natural language processing tasks. Modern state-of-the-art LLMs, such as OpenAI's GPT-4 \cite{openai_gpt-4_2024} or Anthropic's Claude 3.5 Sonnet, have consistently outperformed traditional NLP models in natural language understanding, classification tasks, and information retrieval. These models enable efficient processing of large corpora, reducing the time needed to annotate thousands of documents from weeks or months to just hours \cite{chang2023survey, zhao2023survey, bubeck2023sparks}.

Unlike traditional NLP tools, which often rely on predefined models like sentiment analysis or topic modeling, LLMs can understand nuanced, contextual relationships in text. Previous approaches to analyzing economic narratives frequently employed machine learning pipelines \cite{ash2024relatio}, topic modeling \cite{macaulay2023news}, or sentiment analysis \cite{tilly_predicting_2021}. While these methods are effective for identifying broad patterns, as discussed earlier, they lack the depth to identify and categorize predefined, backward-looking narratives, particularly when such narratives involve subtle linguistic cues or complex causal relationships. On the other hand, studies in computational social sciences have shown that LLMs can match or even surpass human coders in annotating political, social, and economic texts \cite{ziems2024can}, underscoring their potential for content analysis. For example, \citet{mellon2024ais} reported that LLMs achieved 95\% agreement with expert annotators when analyzing British election statements. Similarly, Gilardi et al. \cite{Gilardi_2023} demonstrated that LLMs like GPT-3.5 could classify tweet content, author stances, and narrative frames more accurately than trained crowd workers. Building on this work, we explore the ability of LLMs to identify narratives and narrative-like structures in text. Detailed prompting and the continuous involvement of humans in the loop ensure that LLM annotations align as closely as possible with human intuition.




    

\section{Methodology}
To extract change points in our documents from our corpus, we use the Topical Changes method \citep{TopicalChanges, lange_zeitenwenden_2022}, which is based on the models RollingLDA \citep{rieger_rollinglda_2021} and LDAPrototype \citep{riegerLDAPrototypeModelSelection2024}, that improve the reliability of the classical Latent Dirichlet Allocation \citep{blei_latent_2003} and allows us to apply it to temporal data. After extracting the change points, we further analyze our documents using the LLM Llama 3.1 8B \citep{dubey_llama_2024}. In this section, we detail the models' functionalities and advantages for the task at hand.

\subsection{LDAPrototype}
Because the modeling of LDAs is inherently non-deterministic due to its sampling and initialization, there is no way of telling if a single given run does well to represent the corpus or whether it creates \enquote{bad} topics by chance. To prevent relying on randomness, we use LDAPrototype. 

In terms of language modeling, LDAPrototype follows an LDA \cite{blei_latent_2003}, but instead of training just one LDA, $N$ LDAs are trained from which one model is chosen as a prototypical LDA. To \enquote{represent} the $N$ models, the model with the highest average pairwise similarity to every other model is chosen as the prototype. To do this, \citet{riegerLDAPrototypeModelSelection2024} proposed a similarity measure to compare LDAs with. All pair-wise combinations of $N$ LDA models are compared by clustering their topics based on the cosine similarity of the topic's top words into clusters of size two. The topic model similarity between models $A_1$ and $A_2$ is then given by 
\[\frac{\#\text{topics of model $A_1$ that are matched with a topic of model $A_2$}}{K}\] 
with $K$ as the number of topics in both models $A_1$ and $A_2$. This similarity can thus be interpreted as the percentage of topics that is \enquote{more similar} to a topic of the different model rather than to a topic of the same model, so a form of topic overlap between two models. The chosen prototype will thus be one of the models with the highest average topic overlap and, conversely, the least \enquote{unique} topics that have not been generated by other LDAs and can therefore be considered the most stable out of the $N$ generated models.

\subsection{RollingLDA}
RollingLDA \cite{rieger_rollinglda_2021} is a dynamic topic model based on LDA and can, by extension, also be used with LDAPrototype as a back-end. The model uses a rolling window approach that does not train an LDA from all time chunks at once, but rather begins to model the first $w$ time chunks and then proceeds to model the remaining time chunks based on the information and topic assignments of the last $m$ time chunks. 

As the training of RollingLDA is thus based on the initially trained model, we use $w$ instead of just one time chunk as a warm-up to ensure the model that initializes RollingLDA is properly trained. This parameter should thus be chosen sufficiently large that a proper topic model that covers most important repeating topics. For instance, choosing $w=12$ when using monthly chunks, enables the model to initialize on the first $12$ months of the data without a temporal component, ensuring its initial model has been trained on observing topics that return in yearly trends. 

The memory parameter $m$ creates the rolling window effect of the model. In each time chunk, we provide our model with previous topic assignments from the last $m$ time chunks, which are considered while estimating the document-topic and word-topic distributions during the Gibbs-sampling step. This enables RollingLDA to efficiently model trends in temporal copropra, if the memory parameter is tuned accordingly. For instance, when using monthly chunks, $m=4$ or $m=3$ can be chosen to generate topics \enquote{remembering} quarterly trends, while \enquote{forgetting} older information from years prior. Only providing the model with a bit of past information allows it to change flexibly while still forcing it to keep coherent topics over time. This model is specifically designed to model abrupt changes in rapidly changing news media, which sets it apart from other dynamic topic models, such as the original dynamic topic model \cite{blei_dynamic_2006} and other early iteration of the idea \cite{song_modeling_2005,wang_topics_2006,wang_continuous_2008}. 
\subsection{Topical Changes}
The topical changes model \cite{TopicalChanges, lange_zeitenwenden_2022} detects change points within topics put out by RollingLDA by comparing the development of word-topic vectors over time. For this, we observe both the current and previous count vectors of word assignments and compare the resulting similarity scores. We then perform a Bootstrap-based monitoring procedure that tests whether a change occurred.

To construct these word count vectors, we count the number of occurrences of each word in that topic over the last $z$ time chunks (or since the last detected change, if it occurred less than $z$ time chunks ago). Thus, $z$ is the maximum number of time chunks the change detection can \enquote{look back} to, enabling a rolling window based change detection similar to RollingLDA's rolling window based topic modeling. $z$ can be tuned to, for instance, stop the detection from capturing repeatedly appearing trend effects by setting it as the assumed length of the trend. As the model focuses on finding abrupt changes rather than the slow natural development of language over time, a \enquote{mixture} parameter controls, how much language change the user expects from one time chunk to the next, which alters the look-back word topic vectors by mixing them with the current word vector to a certain degree.

The change detection is then performed by sequentially performing a bootstrap-based test with a significance level $\alpha$ for each time chunk in each topic. In this test, the cosine similarities between the word topic distribution in the current time chunks and $B$ Bootstrap samples of the look-back word topic vector are compared. If the cosine distance of the observed look-back word topic vector to the current word topic vector is larger than $B * (1-\alpha)$ of the bootstrap samples, a change is detected. 

This detection does however not give indication about which tokens have actually caused the change. \citet{TopicalChanges} propose to use words with high leave-one-out word impacts, that is words for which the cosine distance is reduced the most when leaving the word out of both word-topic vectors that are compared during the detection step. These words can thus be interpreted as the main causes for the detected change, as they had the highest impact on the drop of similarity.
\subsection{Llama as a change interpreter}
With the topical changes model, we are able to identify not only points in time where a change point in the topics of our corpus occurred, but also which topics are affected by this change. However, as topics are abstract constructs consisting of word distributions, the model can only give us abstract information about what changed. To gather more information, we further analyze our documents using a large language instruction model with great language understanding and summarization abilities. As we are handling copyrighted texts, we will not use a commercial model such as GPT \citep{openai_gpt-4_2024}, but rather an open source model that runs on a local machine. We use Llama 3.1 8B model, a instruction model designed by Meta with 8 billion parameters \cite{dubey_llama_2024}.

We first narrow down the number of potential documents to give our LLM as an input. Given a change point in topic $k$ between the time chunks $t$ and $t-1$, we employ a filtering strategy that looks for change-related documents to feed into our LLM.

For this, we use the leave-one-out word impacts native to the Topical Changes model as a foundation. We count the number of occurrences of the words found to be significant due to the leave-one-out measure in each document in time chunk $t$. We then select those $5$ documents with the highest count of these words and feed them into our model, letting it compare them to the topic of the previous time chunk, judging from $10$ top words. We do not include documents from time chunk $t-1$ here, as preliminary experiments showed that this \enquote{confused} the model. Leveraging from the Narrative Policy Framework, we give our LLM the following instruction, alongside the $5$ chosen documents:
\begin{quote}
    \#\# You are an expert journalist. You will be asked to explain, why a topical change in a corpus of news articles has has been found and what the change consists of. To fulfill this task, you will be provided information from other text analysis models such as parts of the output of a RollingLDA topic model. \\
    \#\# Whenever you are asked to analyze a \enquote{narrative}, assume the definition of a narrative that is laid out in the paper \enquote{The Narrative Policy Framework: A Traveler’s Guide to Policy Stories}. Specifically, respect and apply the following definitory aspects of a narrative: \enquote{The NPF posits that while the content of narratives may vary across contexts, structural elements are generalizable. For example, the content of a story about fracking told by a Scottish environmentalist is certainly different from the story told by a right-wing populist who attacks a public agency in Switzerland. However, these stories share common structural elements: They take place in a setting, contain characters, have a plot, and often champion a moral.} Keep in mind that a moral must feature a value judgement. When asked to specifiy a moral of a narratives, you must refer to this value judgement or note that there is no moral and thus no narrative! A narrative change must satisfy the four structural criteria, while a content change can simply be caused by an event that shifts the focus of the topic without a clear narrative. Your goal is to determine if a narrative change occurred or if it was a mere content change.\\
    \#\# Please explain an apparent change within a RollingLDA topic that has occurred in [date] \\
    \#\# The following topic top words might give you an idea of what the topic was about before the change: [10 top words of the topic in chunk $t-1$] \\
    \#\# The following topic top words might give you an idea of what the topic was about after the change: [10 top words of the topic in chunk $t$] \\
    \#\# The following words were found to be significant to the detected change: [leave-one-out word impacts]\\
    \#\# The following are those articles from the period that make the most use of the words found to be significant to the detected change: [Filtered articles]\\
    \#\# Provide your output in a strict JSON format. First, summarize each article in one sentence: \{\enquote{summaries}: [\{\enquote{article\_1}: ...\}, \{\enquote{article\_2}: ...\}, ...]\}. Then formulate what the topic was about before and after the change based on the topic top words, emphasizing the changes induced to the topic, judged by the articles and the change words: \enquote{topic\_change}: ... Explain how this change in topic indicates a shift in narrative. How did the narrative shift? \enquote{narrative\_before}: \enquote{Before the change, the narrative centered around ...}, \enquote{narrative\_after}: \enquote{After the change, the narrative centers around ...}. Finally, walk through the four structural criteria that true narratives must satisfy according to the Narrative Policy Framework and confirm or disconfirm their existence in the narrative after the break by briefly naming what they are in the texts provided \{\enquote{narrative\_criteria}: [\{\enquote{setting}: ...\}, \{\enquote{characters}: ...\}, \{\enquote{plot}: ...\}, \{\enquote{moral}: ...\}]\}. Make sure to specify the exact source of the moral judgement that you may have found. Lastly, make a final judgement if there is a narrative shift to be found with \{\enquote{true narrative}: True/False\}. Do not answer in anything but JSON.
\end{quote}
This filtering strategy enables us to specifically observe documents that are found by the model to be significant to the change. It does, however, not capture a larger picture of the topic itself and might lead the model to focus too much on a few significant words. This can happen if more than a few words are significant to the change with similar intensity, not just those that were captured by the leave-one-out word impacts. This might indicate an abrupt and broad topic change that shook the entire word topic distribution.

We also tested out different filtering strategies, such as providing the LLM with documents that are particularly representative of the topic $k$ (i.e. documents with the highest topic share of $k$) in both time frames $t-1$ and $t$, but they generally yielded worse results. We aim to further improve our prompting to further optimize our pipeline in the future.
\section{Evaluation}
We use Python 3.9 to run our scripts. We will publish the code corresponding to this paper as a part of the \href{https://github.com/K-RLange/ttta}{Tools for Temporal Text Analysis (ttta) Python package}\cite{ttta}.

We evaluate our model on a The Wall Street Journal data set containing $795,800$ articles dating from 01/01/2009 to 12/31/2023. This high count of documents allows us to use small time chunks for our RollingLDA analysis. We choose monthly time chunks to enable a fine-grained analysis while ensuring that the number of detected changes remains manageable for annotators. We also choose our memory to check the last four months $m=4$, thus enabling RollingLDA to remember quarterly trends in the data., To provide a stable initial LDA model and to incorporate all yearly trends into that initial model, we choose a warm-up period of $w=12$ months. As a trade-off between computational efficiency and model reliability, we conduct 10 LDA runs to determine an LDAPrototype. We generated the initial model multiple times with $\lbrace 20, 30, ..., 100\rbrace$ topics and decided to choose $K=50$ topics. Our Topical Changes model is then performed with a look-back-window of $4$ months (also to remember for quarterly trends). We use a mixture parameter of $95\%$ and evaluate the similarities to a significance level of $\alpha=0.01$ to control the severity of changes that are observed to be major changes. We choose  $B=500$ Bootstrap samples to generate the bootstrap percentiles. For our LLM we set the temperature parameter to $0$ in the hopes of minimizing hallucinations.

\begin{figure}[t]
\centering
\includegraphics[width=\textwidth]{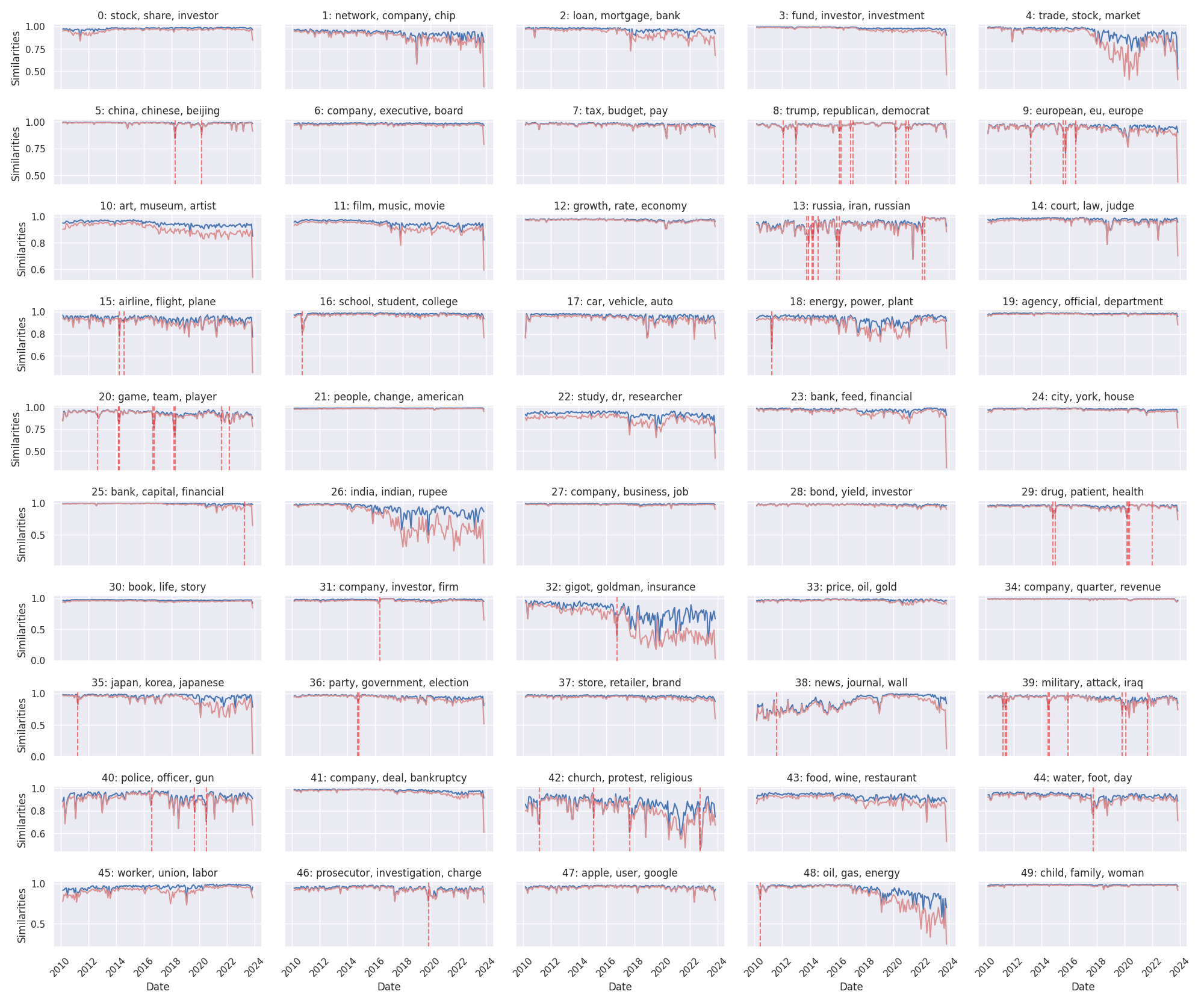}
\caption{Results of the topical changes model with 50 topics on The Wall Street Journal corpus from 2010 to 2023. Vertical red lines indicate a detected change in the topic at that time chunk.}
\label{fig_topical}
\end{figure}

Our Topical Changes model found a total of 68 changes across 156 time chunks. In \autoref{fig_topical} each topic is represented with its overall top words across the entire corpus in its title for interpretability. Each line plot includes a blue line, representing the word vector similarity, going from one chunk to the next while the orange line signalizes the dynamic threshold calculated using bootstrap samples. A change is detected when the blue line falls below the orange line, resulting in a vertical red line. At these change points, we filter the documents for candidate documents to feed into our LLM to check for a narrative shift.

Three expert annotators discussed each change and annotated them according to the Narrative Policy Framework and afterwards classified the answers of the Large Language Model. They found 37 of them to contain narrative rather than mere content shifts. To exemplify Llama's performance, we show three samples of detected changes as well as their narrative shift evaluation of the LLM in the appendix. We will release the list of annotated narratives upon publication of this paper in \href{https://github.com/K-RLange/T2SNarrativeChanges}{a GitHub repository}. 

We split our evaluation into two parts: We see the binary classification between a narrative shift and a content shift as a first step. In the second step, we check if Llama correctly categorized aspects of the existing narratives according to the NPF. The results show that the LLM does not perform particularly well in classifying the changes into content shifts and narrative shifts, as it finds a narrative in 60, so all but eight cases. This results in a accuracy score of 57.35\%, and an f1 score of 0.7010. Since the task involves Llama reproducing all 4 aspects of narratives laid out by NPF, this misclassification is likely caused by hallucinatory behavior, which is a well-known tendency of LLMs \cite{Ji2023}. The LLM hallucinates trying to give a satisfactory output -- ie. provide the user with aspects of narratives according to the NPF  -- at all times, resulting in a large false positive rate.

The model does, however, perform better when explaining an existing narrative shift. In those cases the model accurately defines the narrative in 31 out of 37, so 83,78\%, of the cases. 

Overall, the LLM performs well when a narrative shift exists for a given change point, accurately applying the NPF definition to capture the narrative. However, it does not perform as well when a mere content change occurs, stretching the definitional aspects of the NPF, thus hallucinating narratives. An improved prompt or an additional filtering step could help to solve this issue in future research.

\section{Summary}
When interpreting political or economic events, people align the corresponding information with their internal world view, combining the two into a narrative. Media narratives have become a big research topic in recent years due to the rise of spreaders of \enquote{simple}, often populist narratives. While recent advances in Natural Language Processing, namely the emergence of LLMs, have resulted in improvements in the task of narrative extraction from texts due to their language understanding capabilities, these models are resource intensive. Thus, using them to label narrative in large corpora is often not always feasible.

We introduce a novel pipeline that combines the scalability of the dynamic topic model RollingLDA and its extension Topical Changes with the language understanding of the LLM Llama 3.1 8B. The topic model is used to detect changes within a corpus of The Wall Street Journal dating from 2009 to 2023 in $50$ topics. We use two filtering strategies to identify documents that contain information about the nature of the detected change or the before-and-after of the topic. These texts are then fed to Llama to analyze the change according to the Narrative Policy Framework and detect whether a mere shift in content or in the narrative took place.

We processed the articles in monthly time chunks and detected $68$ changes within 13 years, as one year served as a warm up period. After manually labeling the changes, we find that 37 of those 68 changes show signs of narrative shifts. While our LLM managed to distinguish content shifts from narrative shifts only 57.35\% of the time due to hallucinatory behavior, it correctly explained a narrative shift, if one exists, 83.78\% of the times. The incorrect narrative shift detections stem from the model showing a preference to report a narrative shift for each input rather than considering the option of a change that is not caused by a narrative shift. While this result might improve with more careful prompting, it indicates that the language understanding capabilities of an LLM are functional enough to properly fit an existing narrative shift into a given complex definition, yet may fail not because of insufficient language understanding, but rather due to an inclination to oversatisfy the user’s prompt.

While we observed narratives as purely binary cases in this paper, we plan to perform a more nuanced evaluation of narrative extraction techniques in the future, considering a wider array of subjectivity at this complex task.
\section*{Limitations}
While we demonstrate an exemplary usage of our pipeline here, parameter choices such as the size of time chunks for RollingLDA and the mixture parameter for the Topical Changes can impact the number and granularity of changes detected, which could also alter the results we observed in our study. Furthermore, we didn't observe cases in which the Topical Changes model incorrectly detected a change, we therefore cannot tell, how the LLM would react to such a case.

\section*{Acknowledgments}
This study is part of a project of the Dortmund Center for data-based Media Analysis (DoCMA) at TU Dortmund University. The work was also part of the Narrative Economic Alliance Ruhr (NEAR) project, supported by the Mercator Research Center Ruhr (MERCUR) with project number Ko-2022-0015. 
\bibliography{references2}

\appendix
\section*{Appendix}
\label{appendix}
\subsection*{Sample 1}
\textbf{Topic}: 29 (healthcare)\\
\textbf{Date}: 10/2014\\
\textbf{Leave-one-out word impacts}: ebola, drug, africa, worker, dallas\\
\textbf{Narrative shift?}: Yes\\
\textbf{Hand-coded narrative shift}: After a small ebola outbreak in the U.S. in the month prior and an appropriate response that condemned that outbreak, the focus of the topic shifts. Instead of the fear of an outbreak in the U.S., the articles focus on drug tests that give hope and also shift the focus back on the African continent.\\
{\large\textbf{Response of the LLM}:}\\
\{\\
\textbf{summaries}: [\\
    \{article\_1: \enquote{Pharmaceutical companies and health officials are racing to develop and test treatments for the Ebola virus, with several experimental drugs showing promise in early trials.}\},\\
    \{article\_2: \enquote{The US response to the Ebola outbreak has improved significantly since the first case in Dallas, with officials learning from past mistakes and taking steps to contain the virus.}\},\\
    \{article\_3: \enquote{The number of people potentially exposed to Ebola in the US has dropped to 50 from 100, and officials are working to reassure the public that the virus can be contained.}\},\\
    \{article\_4: \enquote{A nurse in Texas has tested positive for Ebola, highlighting the need for hospitals to be more vigilant in their infection-control procedures and raising concerns about the preparedness of US hospitals for the disease.}\},\\
    \{article\_5: \enquote{The Ebola outbreak in West Africa is being hindered by a lack of qualified staff, with many health workers in Liberia and other affected countries lacking proper training and equipment to combat the virus.}\}
  ],\\
  \textbf{topic\_change}: \enquote{The topic shifted from a focus on the global response to the Ebola outbreak, including the development of treatments and the US response, to a focus on the challenges of containing the outbreak in West Africa, particularly the lack of qualified staff and the need for improved infection-control procedures.},\\
  \textbf{narrative\_before}: \enquote{Before the change, the narrative centered around the global response to the Ebola outbreak, with a focus on the development of treatments and the US response, emphasizing the sense of urgency and the need for action.},\\
  \textbf{narrative\_after}: \enquote{After the change, the narrative centers around the challenges of containing the outbreak in West Africa, highlighting the difficulties of providing adequate care and the need for improved infection-control procedures, with a focus on the human cost of the outbreak and the need for more effective solutions.},\\
  \textbf{narrative\_criteria}: [\\
    \{\textbf{setting}: \enquote{The setting of the narrative is the Ebola outbreak in West Africa, particularly in Liberia, where the lack of qualified staff and inadequate infection-control procedures are exacerbating the crisis.}\},\\
    \{\textbf{characters}: \enquote{The characters in the narrative include health workers, patients, and officials, who are struggling to contain the outbreak and provide adequate care.}\},\\
    \{\textbf{plot}: \enquote{The plot of the narrative revolves around the challenges of containing the outbreak, including the lack of qualified staff, inadequate infection-control procedures, and the need for improved solutions.}\},\\
    \{\textbf{moral}: \enquote{The moral of the narrative is that the Ebola outbreak highlights the need for improved global health infrastructure and the importance of prioritizing the health and safety of healthcare workers, as emphasized by Dr. Frieden, the CDC Director, who stated that \enquote{preventing transmission requires scrupulous attention to infection control}.}\}\\
  ],\\  
  \textbf{true\_narrative}: true
\}\\
\textbf{LLM evaluation}: LLM detected the fact that a narrative shift correctly. It correctly identified that the focus shifted back to the African continent and condemning the disease there. It did not focus on the emotional aspect of \enquote{hope vs. fear}, which might be the result of not having been provided from the month prior to the change, resulting in a lack of context.

\subsection*{Sample 2}
\textbf{Topic}: 36 (Elections and Polls)\\
\textbf{Date}: 10/2014\\
\textbf{Leave-one-out word impacts}: scotland, scottish, independence, neves, scot\\
\textbf{Narrative shift?}: No\\
\textbf{Hand-coded narrative shift}: The protagonists (and settings) in this topic have hardly anything in common. Documents deal with Catalonia's independence movement, Brazil's presidential election, and Scotland's decision to remain in the U.K., among others. There is no such thing as a common narrative here.\\
{\large\textbf{Response of the LLM}:}\\
\{
\textbf{summaries}: [\\
    \{article\_1: \enquote{Catalonia's independence movement remains strong despite a court injunction, with activists planning to hold a nonbinding referendum on November 9.}\},\\
    \{article\_2: \enquote{Aécio Neves, the opposition candidate in Brazil's presidential election, is gaining momentum against incumbent President Dilma Rousseff, with polls showing him within striking distance of a runoff.}\},\\
    \{article\_3: \enquote{Marina Silva, the third-place finisher in Brazil's presidential election, is considering endorsing Aécio Neves, the opposition candidate, in a bid to defeat President Dilma Rousseff in a runoff.}\},\\
    \{article\_4: \enquote{The Brazilian presidential election is too close to call, with Aécio Neves and Dilma Rousseff neck and neck in the polls, with Neves focusing on the economy and Rousseff maintaining a loyal base of support.}\},\\
    \{article\_5: \enquote{The Bank of England had planned to flood the financial system with cash and hold emergency auctions to secure banks' funding if Scotland had voted for independence in its referendum.}\}
  ],\\
  \textbf{topic\_change}: \enquote{The topic shifted from a discussion of Scottish independence to a discussion of Brazilian politics and the presidential election, with a focus on the opposition candidate Aécio Neves and his chances of defeating incumbent President Dilma Rousseff.},\\
  \textbf{narrative\_before}: \enquote{Before the change, the narrative centered around the Scottish independence movement, with a focus on the potential consequences of a 'yes' vote and the plans of the Bank of England to maintain financial stability.},\\
  \textbf{narrative\_after}: \enquote{After the change, the narrative centers around the Brazilian presidential election, with a focus on the opposition candidate Aécio Neves and his chances of defeating incumbent President Dilma Rousseff.},\\
  \textbf{narrative\_criteria}: [\\
    \{\textbf{setting}: \enquote{The setting is the countries of Scotland and Brazil, with a focus on the political and economic systems of each country.}\},\\    
    \{\textbf{characters}: \enquote{The main characters are the politicians and activists involved in the Scottish independence movement and the Brazilian presidential election, including Aécio Neves, Dilma Rousseff, and Marina Silva.}\},\\
    \{\textbf{plot}: \enquote{The plot revolves around the potential consequences of a 'yes' vote in the Scottish referendum and the outcome of the Brazilian presidential election, with a focus on the opposition candidate Aécio Neves and his chances of defeating incumbent President Dilma Rousseff.}\},\\
    \{\textbf{moral}: \enquote{There is no clear moral judgment in the narrative, but the narrative does suggest that the opposition candidate Aécio Neves is a more viable option for Brazil's economy and that the incumbent President Dilma Rousseff's policies have been ineffective in addressing the country's economic problems.}\}]\\
  \textbf{true\_narrative}: true
\}\\
\textbf{LLM evaluation}: The model correctly detected some of the main characters. However, it did not come to the conclusion that, due to the different settings, the provided documents are more or less incomparable. The suggested moral may fit to a single document, but does certainly not consider all provided texts adequately.

\subsection*{Sample 3}
\textbf{Topic}: 8 (US Elections)\\
\textbf{Date}: 01/2016\\
\textbf{Leave-one-out word impacts}: trump, house, iowa, cruz, sander\\
\textbf{Narrative shift?}: Yes\\
\textbf{Hand-coded narrative shift}: Just before the start of the 2016 Republican primaries, Donald Trump's campaign has gained momentum, overtaking his more mainstream competitors through his charisma and aggressive attacks on competitors, such as Ted Cruz and Marco Rubio, and despite questions surrounding his character and previous business dealings.\\
{\large\textbf{Response of the LLM}:}\\
\{
\textbf{summaries}: [\\
    \{article\_1: \enquote{The article discusses Donald Trump's financial struggles in the 1990s, including his debt restructuring and bankruptcy, and how he managed to come out of it by taking advantage of his creditors and using his charisma to negotiate deals.}\},\\
    \{article\_2: \enquote{The article discusses the 2016 presidential election and how Donald Trump's campaign is gaining momentum, despite his high disapproval ratings and the objections of some voters, and how he may be able to overcome these obstacles by confronting his opponents and using his charisma to win over voters.}\},\\
    \{article\_3: \enquote{The article discusses the 2016 presidential election and how Ted Cruz and Donald Trump are battling for the lead in Iowa, with Cruz's campaign accusing Trump of being too egotistical and Trump questioning Cruz's eligibility to be president.}\},\\
    \{article\_4: \enquote{The article discusses the 2016 presidential election and how Donald Trump is targeting Ted Cruz, questioning his eligibility to be president and his evangelical faith, and how Cruz is handling the attacks with humor and poise.}\},\\
    \{article\_5: \enquote{The article discusses the final debate before the Iowa caucuses, where the candidates, minus Donald Trump, sparred over policy differences and personal attacks, with Ted Cruz and Marco Rubio engaging in a heated exchange over immigration policy.}\}
  ],\\
  \textbf{topic\_change}: \enquote{The topic shifted from discussing Donald Trump's financial struggles and personal characteristics to focusing on his presidential campaign and the 2016 election, with a particular emphasis on his rivalry with Ted Cruz and the other Republican candidates.},\\
  \textbf{narrative\_before}: \enquote{Before the change, the narrative centered around Donald Trump's personal characteristics, such as his financial struggles, his ability to negotiate deals, and his charisma, and how these traits might make him a successful businessman and president.},\\
  \textbf{narrative\_after}: \enquote{After the change, the narrative centers around Trump's presidential campaign and his rivalry with other candidates, particularly Ted Cruz, and how he is using his charisma and tactics to win over voters and gain momentum in the election.},\\
  \textbf{narrative\_criteria}: [\\
    \{\textbf{setting}: \enquote{The setting of the narrative is the 2016 presidential election, with the context of the Republican primary and the Iowa caucuses.}\},\\
    \{\textbf{characters}: \enquote{The main characters in the narrative are Donald Trump, Ted Cruz, and the other Republican candidates, who are portrayed as rivals and opponents in the election.}\},\\
    \{\textbf{plot}: \enquote{The plot of the narrative is the struggle for the Republican nomination, with Trump and Cruz engaging in a heated rivalry and the other candidates trying to position themselves as alternatives to the two front-runners.}\},\\    
    \{\textbf{moral}: \enquote{There is no clear moral judgment in the narrative, but the story implies that Trump's ability to win over voters and gain momentum in the election is a testament to his charisma and tactics, rather than any inherent qualities or values.}\}
  ]\\
  \textbf{true\_narrative}: true
\}\\
\textbf{LLM evaluation}: 
The model correctly identified the main characters, the setting and the plot of Donald Trump taking the leading role in the primary field. While the model somewhat overstated the importance of the \enquote{moral} implications of this development, all aspects of the NPF were correctly recognized and summarized.
\end{document}